# Kutató: An Entropy-Driven System for Construction of Probabilistic Expert Systems from Databases

Edward Herskovits, M. D.
Gregory Cooper, M. D., Ph. D.

Knowledge Systems Laboratory
Medical Computer Science
Stanford University

## Abstract

Kutató[1] is a system that takes as input a database of cases and produces a belief network that captures many of the dependence relations represented by those data. This system incorporates a module for determining the entropy of a belief network and a module for constructing belief networks based on entropy calculations. Kutató constructs an initial belief network in which all variables in the database are assumed to be marginally independent. The entropy of this belief network is calculated, and that arc is added that minimizes the entropy of the resulting belief network. Conditional probabilities for an arc are obtained directly from the database. This process continues until an entropy-based threshold is reached. We have tested the system by generating databases from networks using the probabilistic logic-sampling method, and then using those databases as input to Kutató. The system consistently reproduces the original belief networks with high fidelity.

## 1.    Introduction

Computer-based information processing has changed dramatically since the construction of the first computers. From its inception as an arithmetic discipline, the field has evolved to provide sophisticated means for increasing our understanding of nature. This evolution has been made possible by the availability of increasingly sophisticated hardware and software, and has been driven by the rapid growth of information from experiments. Data are being generated so rapidly in some fields that manual or even semiautomated methods of data analysis cannot keep pace, resulting in databases that remain unexplored.

---

[1]    Kutató means "explorer" or "investigator" in Hungarian.

Researchers have explored databases for several reasons, most notably to discover and to validate knowledge [Walker, 1990]; here, we focus on the automated or semiautomated construction of probabilistic expert systems, a form of knowledge discovery. In particular, we address the problem of constructing a *Bayesian belief network*, herein referred to as a belief network, from a database. We direct the reader to [Cooper, 1989; Horvitz, 1988] for introductions to belief networks and their relation to other expert-system paradigms; to [Ross, 1984] and to [Jaynes, 1982; Levine, 1978] for introductions to the concepts of entropy and maximum entropy, respectively; to [Cohen, 1982; Michalski, 1983; Michalski, 1986] for discussions of machine learning based on artificial-intelligence techniques; to [Glymour, 1987] for an analysis of the determination of causal structure based on statistical methods; and to [Breiman, 1984] for a discussion of discovering associations among variables by recursive partitioning of a data set.

Many of the numerical algorithms for database exploration have their roots in information theory; in particular, they share a foundation on the principle of maximum entropy [Jaynes, 1982]; the entropy of a distribution is calculated using the equation

$$H = -\sum P_i \, log \, P_i .$$

where $P_i$ corresponds to an element in the full joint distribution (there would be $2^n$ terms for a distribution based on $n$ binary variables). The maximum-entropy principle is invoked when there is insufficient information to determine the full joint distribution unambiguously. The principle states that, in the absence of prior information about the distribution, by choosing the full joint distribution that has maximum entropy given the information at hand, we guarantee that probabilities derived from the resulting distribution will have no bias. This result is unique to the principle of maximum entropy; imposing any other constraints (such as that of a particular distribution class) on the data may introduce biases.

From the perspective of reconstructing a probability distribution from a database, researchers have employed the principle of maximum entropy by treating the cases in a database as constraints on an underlying distribution; since most databases have far fewer cases than elements in the corresponding full joint distribution, the latter is severely underconstrained. Algorithms using this principle return a list of probabilities from the database that, taken together, represent all the interdependence in the underlying full joint distribution; that is, any probabilities in the full joint distribution not in the list may be calculated from those in the list, since the list is assumed to capture all significant dependencies among variables. For example, if the list consisted solely of first-order probabilities, all higher-order probabilities could be calculated as products of these first-order probabilities. In addition to its use in estimating distributions, entropy calculation has been used to perform probabilistic inference [Wen, 1988] and to generate production rules from a database [Chan, 1989; Cheeseman, 1983; Gevarter, 1986; Goodman, 1989].

Researchers first used entropy-based methods for database exploration as a byproduct of investigating the more general problem of generating a parsimonious probability distribution that best approximates a known underlying distribution. Lewis [Lewis II, 1959] described an algorithm for approximating an $n$th-order binary distribution as a product of lower-order distributions, based on the *closeness metric*:

$$I_{P\text{-}P'} = \sum_{j=0}^{2^n-1} P_j ln \frac{P_j}{P'_j},$$

which is also known as the Kullback–Liebler cross-entropy measure [Kullback, 1951]. This number is 0 if and only if the two distributions $P$ and $P'$ are identical; otherwise, it is positive. The algorithm searches a strongly restricted subset of possible approximating distributions $P'$ (those that have the same lower-order joint probabilities as those of the true distribution $P$), and chooses the distribution that minimizes the closeness metric.

Chow and Liu [Chow, 1968] considered the approximation of an $n$th-order distribution with $n - 1$ second-order distributions, using Lewis' closeness metric. In this algorithm only $\binom{n}{2}$ numbers need to be determined; they correspond to pairwise associations, and are added incrementally until the $n - 1$ strongest have been included, at which time the program terminates. Although this algorithm is relatively efficient computationally, it is highly restricted in that only those approximating distributions composed of second-order probabilities are considered.

Ku and Kullback [Ku, 1969] generalized Chow and Liu's algorithm, allowing any lower-order marginal distributions to be used in approximating an $n$th-order distribution. A convergent iterative formula is used to determine the distribution given a set of lower-order marginal constraints. As expected, the algorithm converges on increasingly accurate approximations as the order of the marginal distributions is increased; this accuracy is obtained at the cost of running times and data requirements that are exponential in the order of the approximating distribution, as each element of that distribution must be estimated with the convergent iterative algorithm.

An algorithm developed by Cheeseman [Cheeseman, 1983] and augmented by Gevarter [Gevarter, 1986] maximizes the entropy of a distribution given only first-order constraints obtained from data. The algorithm searches for significant constraints heuristically, in contrast to the iterative, exhaustive methods used by previous workers. A significance test is employed to determine whether the second-order probabilities derived from the database are significantly different from those obtained from the maximum-entropy distribution. If they are, the most extreme deviant (itself a second-order probability) is added to the set of constraints, and an enhanced maximum-entropy distribution is computed. This process continues until no further second-order constraints are significant; the algorithm then continues to test third- and higher-order constraints until there remain no statistically significant differences between the distribution computed from maximum-entropy considerations and that derived from the database. In summary, this procedure represents a myopic search progressing from the lowest-order constraints to the highest-order constraints embodied in the database, and uses a significance test at each step to determine whether any probabilities in the database are different from their expected values given the constraints already found.

Because the general methods of Ku and Kullback and of Cheeseman and Gevarter rely on iterative algorithms and have running times that are exponential in the order of the approximating distribution, researchers sought to bring other computational techniques to bear on the problem of constructing a probabilistic expert system from data. Pearl [Pearl, 1988] discussed the separation of what he called *structure learning,* determining a dependency model for a probability distribution, and *parameter learning,* determining the probabilities that complete that model. From this perspective, the algorithm developed by Chow and Liu returns a *dependence tree,* which is a spanning tree representing all significant pairwise correlations among variables, and a set of second-order probabilities for each (undirected) arc in that tree. Pearl also described a method whereby a polytree could be extracted from a probability distribution using calculations similar to those specified by Chow and Liu in addition to partial structure determinations based on tests of conditional independence. This algorithm, although more general than that described by Chow and Liu, is not guaranteed to find the best polytree-based approximation to an arbitrary distribution; furthermore, the algorithm cannot return nontree structures.

Extending the distinction between structure and parameter learning, Spiegelhalter and Lauritzen [Spiegelhalter, 1989] maintained that structure learning should occur only in the presence of a domain expert, and described a method for using data to update the conditional probabilities in a belief network whose structure has been specified by an expert. The method is based on local- and global-independence assumptions; the former allows the algorithm to individually parameterize each particular conditional-probability distribution for a node given a particular instantiation of its parents, and the latter allows the algorithm to compute the belief network's distribution as the product of the distributions for each node. The authors use a Dirichlet distribution to parameterize the conditional-probability distributions parsimoniously and to provide a basis for locally updating these conditional-probability functions via approximations to the resulting finite-mixture distributions. The strong independence assumptions and updating heuristics allow incremental updating of the conditional probabilities (that is, on a case-by-case basis), at the cost of maintaining the network structure constant over time, and with the restriction that these techniques be applied only to domains that manifest global and local independence.

In contrast to the approach of Spiegelhalter and Lauritzen regarding automated parameter determination, Srinivas, Russell, and Agogino [Srinivas, 1989] posited that a system that can learn structure from data or other constraints might alleviate the knowledge-acquisition bottleneck. They developed an algorithm that takes as input some qualitative information from an expert about the dependencies in the domain, and returns a belief network incorporating these constraints. No attempt is made to use data to compute conditional probabilities; only the structure is determined. The expert-derived information about a variable or set of variables may be stated in any one of four forms:

- A variable $X$ is a root node, or hypothesis variable

- A variable $X$ is a leaf node, or evidence variable
- A variable $X$ is a direct predecessor, or parent, of $Y$ (that is, $X$ causes $Y$)
- Variable sets $X$ and $Y$ are conditionally independent given set $Z$

The algorithm applies a priority heuristic to each node, adding hypotheses, causes, effects, and evidence nodes to the nascent belief network in that order; it breaks ties by adding the node that would bring with it the fewest arcs. This process continues until all nodes have been added to the network. The algorithm's computational complexity is exponential in the number of nodes, does not use data, and does not compute conditional probabilities, although in principle the last two issues could be addressed with extensions to the algorithm.

## 2. The Kutató Algorithm

We have developed an algorithm, called Kutató, that, given a database, returns a belief network. Kutató determines the network's structure by beginning with the assumption of marginal independence among all variables, and by adding the arc that maintains acyclicity and results in a belief network with minimal entropy. We attempt to minimize entropy since we are approaching the maximum-entropy distribution from above. The arc-addition step represents an attempt to find the association that most strongly constrains the ensuing distribution. As an arc is added, the database is used to update the conditional-probability distribution for the node at the head of the new arc. Arcs are added in this manner until a threshold is reached in the rate of decrease of the entropy between two successive networks. Consider an $n$-node model; since any two nodes may in principle be associated, $O(n^2)$ arcs are considered before the best one to add (if any) is chosen; further, since in principle all these associations may be found to be significant, this cycle is repeated $O(n^2)$ times, resulting in a complexity (not including entropy calculations) of $O(n^4)$.

Directions of arcs are dictated by a *total order* on variables in the database, although the alternative of having the algorithm choose arc directions based on entropy mini-mization is also available to the user. Kutató obtains the total order from a domain expert by having him answer the question, "For the two variables $A$ and $B$, which one *cannot* cause the other?" for each pair of variables $(A, B)$ in the database. (If the answer is not known, a random order may be assigned.) This procedure obviates complex reasoning about causality, results in a more intuitively appearing belief network, and provides a relatively simple and efficient method for obtaining rudimentary causal knowledge; yet, it is not required. Indeed, the user might supply an order resulting in a more highly connected network than would have resulted without any causal information. Thus, in some cases, there may be a tradeoff between choosing the directions of the arcs and decreasing the interconnectedness of the final network.

### 2.1 The Entropy-Computation Algorithm

Given that inference in belief networks is NP-hard [Cooper, 1987], it is not surprising to find that the problem of determining the entropy of an arbitrary probability distribution is NP-Hard [Cooper, 1990a]. Just as other workers have exploited the principle of conditional independence to increase greatly the efficiency of inference, we have developed an algorithm that exploits the conditional independence embodied in a belief network to compute its distribution's entropy. Using this algorithm is much more efficient than is summing over the full joint distribution, which makes this project feasible. As discussed in Section 2, in the worst cases, the entropy calculations must be performed $O(n^4)$ times, thus making the overall complexity of Kutató $O(n^4 2^n)$. We emphasize that this upper bound on complexity represents the worst case, wherein each node is directly connected to every other node. We expect many realistic models to be sparsely connected, and indeed, this has been our experience.

The formula for calculating the entropy of a distribution represented by a belief network is based on the concept of conditional entropy [Ross, 1984]. Let $U$ be the set of nodes in a belief network $BN$; for any node $X \in U$, let $\Pi_X$ be the set of its parents (direct predecessors), and let $\pi_x$ be a particular instantia-tion of the parents of $X$. The entropy of the distribution represented by $BN$ is

$$H_{BN} = \sum_{X \in U} \sum_{\pi_X} P(\Pi_X = \pi_X) \, H_{X \mid \pi_X},$$

where

$$H_{X \mid \pi_X} = \sum_x P(X = x \mid \Pi_X = \pi_X) \, \ln P(X = x \mid \Pi_X = \pi_X).$$

These formulas state that we can calculate the entropy of a distribution represented by a belief network by weighting each node's conditional entropy given a particular instantiation of that node's parents by the joint probability of the parents' assuming those values. We implemented a modified version of this formula using Cooper's recursive decomposition algorithm [Cooper, 1990b]. With this implementation, we can compute the entropy of ALARM [Beinlich, 1989]—a belief network with 37 nodes, 46 arcs, and approximately $10^{17}$ elements in its joint distribution—in less than 10 seconds. Conditional independence provides the computational leverage that allows this calculation to be performed efficiently.

## 2.2 The Significance Test

Each cycle of the algorithm yields a set of $O(n^2)$ entropy measures corresponding to the individual additions of each possible remaining arc. A function is needed as a means of determining the best arc to add, or whether the program should halt. We chose to test for significance using the change in entropy of the underlying distribution, because entropy is sensitive to bias, and because we can formulate a straightforward significance test based on entropy changes, as shown in [Jaynes, 1982]. Jaynes demonstrated that the test statistic $2N\Delta H$, where $N$ is the number of cases used to update the network, and $\Delta H$ is the difference in entropy that results from adding an arc to the network, is asymptotically (as $N \to \infty$) distributed as chi-squared. We can use this result in constructing a significance test.

For each arc considered during a cycle of the algorithm, we compute the probability that the distribution represented by the belief network including the arc is the same as the distribution of that network without the arc. Computing the entropy difference between the two networks, we can employ a chi-squared test with the appropriate degrees of freedom. We then have, for each arc, a probability that that arc's addition makes no difference in the underlying distributions; this result corresponds to conditional independence. By choosing the arc with the lowest probability of manifesting conditional independence, we maximize the probability that this arc should be added to the belief network.

## 2.3 The Dirichlet Distribution

Any classification or exploration system must have a method for managing incomplete information. In particular, systems that examine databases for interdependence among variables are plagued by the problem of overfitting the data. For example, a data set could be partitioned into so many elements that each unique case is grouped alone; this result is equivalent to maintaining the full joint distribution, which is unwieldy. Furthermore, in some sense, overclassification can be viewed as an algorithm's overconfidence in how well the data represent the underlying distribution. Most databases have far fewer cases than they have elements in the corresponding full joint distribution, so this distribution is severely underconstrained. Here is another case where the maximum-entropy principle could be employed, yet it is prohibitively expensive computationally to compute the entropy of a database. It would be much more convenient to compute the entropy of a belief network *derived from* a database.

As an alternative, we can consider the database to represent a sample from an infinitely exchangeable multinomial sequence; we can then use symmetric Dirichlet prior probabilities for computing conditional probabilities [Zabell, 1982]. In particular, for node $X$ having $V_X$ values, parents $\Pi_X$, and considering a particular instantiation of those parents $\pi_X$, we compute the corresponding conditional probability with the following formula:

$$P\big(X = x \mid \Pi_X = \pi_X\big) = \frac{C\big(X = x, \Pi_X = \pi_X\big) + 1}{C\big(\Pi_X = \pi_X\big) + V_X},$$

where $C(\Phi)$ is the number of cases that match the instantiated set of variables $\Phi$.

The use of such prior probabilities addresses several problems. When we are attempting to determine the conditional probabilities for an arc that is not represented in the database, the principle of maximum entropy, if applied locally, would generate a uniform distribution for these conditional probabilities; this result also follows when we use the Dirichlet distribution. In addition, this method allows Kutató to handle incomplete data. Only those cases that can update the conditional-probability table are used; if none exist, a default uniform distribution results. Using Dirichlet prior probabilities further results in a natural halt to overspecification: When a uniform conditional-probability distribution is generated (due to an absence of relevant cases), the entropy of the belief network will *rise*, leading to prompt rejection of the corresponding arc. Indeed, this effect is also observed when the number of relevant cases is small, since the resulting distribution will still be approximately uniform.

## 3.   Results[2]

We tested Kutató by acquiring a belief network, generating a database of cases with the probabilistic logic-sampling method [Henrion, 1988], and then using that database as input to Kutató. The first belief network tested was MCBN1, a binary network of five nodes and five arcs (see Figure 1); its full joint distribution thus has 32 elements, and the probabilities in that distribution range from 0.00024 to 0.46656. Because this distribution has few elements, we were able to test the Kutató with the exact full joint distribution, the equivalent of an infinite database, instead of data. Kutató returned MCBN1 exactly, in 13 seconds. We then used logic sampling to generate a database of 1000 complete cases. Kutató generated MCBN1's structure exactly in less than 1 minute (two-thirds of this time was spent reading the database), and all of the conditional probabilities (ranging from 0.1 to 0.9) were accurate to within 0.04.

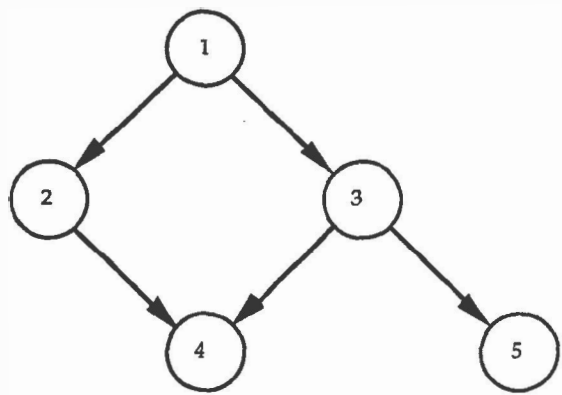

Figure 1   The MCBN1 binary belief network, with five nodes and five arcs.

We next tested the ALARM belief network (see Figure 2), using a database of 10,000 complete cases. The resulting network was generated in approximately 22.5 hours (one-fourth of which was spent reading the database); it is shown in Figure 3. The program added 46 arcs before halting (the original version of ALARM also has 46 arcs). Two arcs of 46 were missing, and two extra arcs were added.

## 4.   Future Research

We will apply Kutató to a series of databases in a variety of domains. We also will investigate the system's behavior in the face of increasingly sparse data. Several other possible extensions to this work include:

- A probabilistic reformulation of this work. One such algorithm, K2, has been developed by Cooper, and is being investigated by the authors. Preliminary results indicate that, compared to Kutató, this algorithm runs faster and is more robust to noise.

- A version of K2 based on continuous distributions. A version based on the multivariate Gaussian distribution would complement Shachter's work on Gaussian influence diagrams [Shachter, 1989].

- Modifications of the greedy search used for arc addition. For example, several arcs could be added at a time, or arcs could be deleted.

- Development of a template for temporal models. Variables could be modeled during several discrete time periods to

---

[2]   The software was implemented and evaluated on a Macintosh II using Lightspeed Pascal v. 2.0.

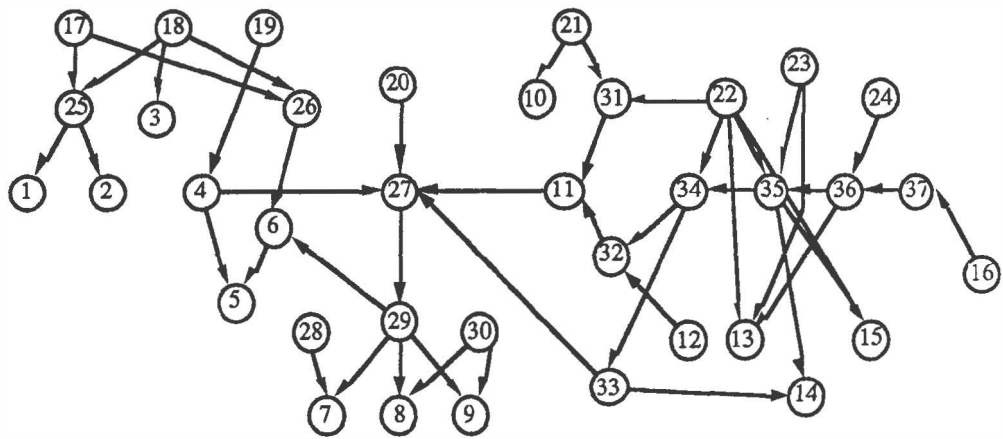

Figure 2 The ALARM belief network, with 37 nodes and 46 arcs.

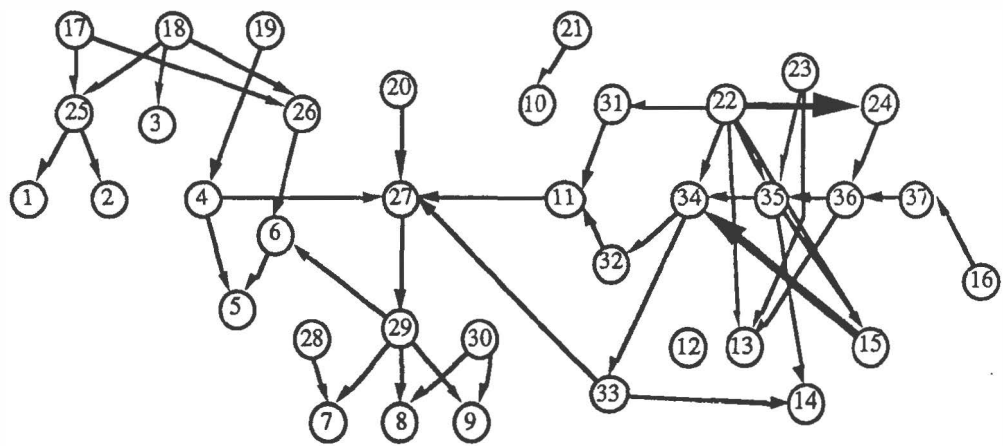

Figure 3 The ALARM network generated by Kutató from a 10,000-case database. The arcs from node 21 to node 31 and from node 12 to node 32 are missing, and extra arcs (bold) from node 15 to node 34 and from node 22 to node 24 have been added.

determine time-lagged probabilistic associations among variables.

- Delineation of a constraint language. It should be capable of expressing expert-derived constraints on relationships among variables in the database. This language would greatly extend the system's expressiveness beyond the current total order used to determine the direction of arcs. K2, unlike Kutató, can incorporate prior probabilities over possible networks; thus, we will apply this language to K2.

## 5. Conclusion

Kutató is an efficient system for approximating the maximum-entropy distribution of a database. It is applicable in the presence of missing data, noisy data (such as those obtained from probabilistic logic-sampling), and immense joint-probability spaces. The algorithm makes use of the conditional independence manifested in a belief network to streamline computation, enabling us to run the program on readily available hardware. A new, probabilistic version of this algorithm show even greater promise for constructing probabilistic networks from data.

## Acknowledgments

We thank Lyn Dupré for helpful comments on an earlier version of this paper. This work was supported by grant LM05208 from the National Library of Medicine, by grant P-25514-EL from the U. S. Army Research Office, and by grant IRI-8703710 from the National Science Foundation. Computer facilities were provided in part by the SUMEX-AIM resource under grant LM05208 from the National Library of Medicine.

# *Session 3:*

## Control of Reasoning in Belief Networks